\documentclass[sigconf]{acmart}
\usepackage{booktabs} % For formal tables

\usepackage{tcolorbox}
\usepackage{tabularx}
\usepackage{array}
\usepackage{colortbl}
\usepackage{comment}
\usepackage{soul} % for \hl highlighting
\tcbuselibrary{skins}

\usepackage{placeins}

\newcolumntype{Y}{>{\raggedleft\arraybackslash}X}

\tcbset{tab1/.style={fonttitle=\bfseries\large,fontupper=\normalsize\sffamily,
colback=white!10!white,colframe=black!75!black,colbacktitle=Salmon!40!white,
coltitle=black,center title,freelance,frame code={
\foreach \n in {north east,north west,south east,south west}
{\path [fill=black!75!black] (interior.\n) circle (0.5mm); };},}}

\tcbset{tab2/.style={enhanced,fonttitle=\bfseries,fontupper=\normalsize\sffamily,
colback=white!10!white,colframe=black!50!black,colbacktitle=Salmon!40!white,
coltitle=black,center title}}

\newcommand{\sys}{\textit{MobText-SISA}}

\usepackage[subrefformat=parens]{subcaption}
\begin{document}

% \title{\sys{}: Efficient Indoor Localization  Based on Single WiFi Access Point And Minimal Fingerprinting }
%\vspace{-2cm}
\title{\textcolor{black}{\sys{}: Efficient Machine Unlearning for Mobility Logs with Spatio-Temporal and Natural-Language Data}}

% \author{Yuya Takeuchi}
% \affiliation{
% \institution{Osaka University, Japan}
%     \city{}
%     \country{}
% }
% % \email{y-takeuchi@ist.osaka-u.ac.jp}

% \author{Haruki Yonekura}
% \affiliation{
% \institution{Osaka University, Japan}
%     \city{}
%     \country{}
% }
% % \email{h-yonekura@ist.osaka-u.ac.jp}

% \author{Kyosuke Yamashita}
% \affiliation{
% \institution{Osaka University, Japan}
%     \city{}
%     \country{}
% }
% % \email{k-yamashita@ist.osaka-u.ac.jp}

% \author{Hirozumi Yamaguchi}
% \affiliation{
%     \institution{Osaka University, RIKEN Center for Computational Science, Japan}
%     \city{}
%     \country{}
% }
% % \email{h-yamagu@ist.osaka-u.ac.jp}

%     \vspace{-2em}
%     \begin{center}
%     \texttt{\{y-takeuchi, h-yonekura, yamashita, h-yamagu\}@ist.osaka-u.ac.jp}
%     \end{center}

% \author{Haruki Yonekura$^{1,2}$, Ren Ozeki$^{1,3}$, Tatsuya Amano$^{1,2}$, Hamada Rizk$^{1,2,4}$, Hirozumi Yamaguchi$^{1,2}$}
\author[Haruki Yonekura, Ren Ozeki, Tatsuya Amano, Hamada Rizk, Hirozumi Yamaguchi]{%
  Haruki Yonekura$^{1,2}$, 
  Ren Ozeki$^{1,3}$, 
  Tatsuya Amano$^{1,2}$, 
  Hamada Rizk$^{1,2,4}$, 
  Hirozumi Yamaguchi$^{1,2}$
}
\def \authors{Haruki Yonekura, Ren Ozeki, Tatsuya Amano, Hamada Rizk, Hirozumi Yamaguchi}
\affiliation{%
\institution{$^1$The University of Osaka
\country{Japan}}
}
\affiliation{%
  \institution{$^2$RIKEN Center for Computational Science
  \country{Japan}}
}
\affiliation{%
  \institution{$^3$Japan Meteorological Agency
  \country{Japan}}
}
\affiliation{%
  \institution{$^4$Tanta University
  \country{Egypt}}
}

\begin{abstract}
% \color{blue}
% AI-driven mobility services now train deep models on massive spatio–temporal logs.  Care-taxi dispatch records store precise pick-up and drop-off coordinates; our dataset further preserves brief operator notes on passenger assistance—an under-exploited textual modality that can reveal health conditions. These logs thus fall under the GDPR right to be forgotten, requiring that a requester’s influence be removed from both storage and model parameters.  Retraining a continuously updated model from scratch for every deletion is infeasible.  We propose \sys{}, an unlearning framework tailored to mobility data that extends Sharded, Isolated, Sliced, and Aggregated (SISA) training.  \sys{} clusters trajectories and their associated text into latent passenger groups, partitions the data accordingly, and trains an ensemble whose members can be retrained in isolation.  Experiments on ten months of real care-taxi data show that \sys{} reduces unlearning cost while maintaining predictive accuracy, enabling practical, privacy-compliant analytics for ageing-society transport.
Modern mobility platforms have stored vast streams of GPS trajectories, temporal metadata, free-form textual notes, and other unstructured data. Privacy statutes such as the GDPR require that any individual’s contribution be unlearned on demand, yet retraining deep models from scratch for every request is untenable. We introduce \sys{}, a scalable machine-unlearning framework that extends Sharded, Isolated, Sliced, and Aggregated (SISA) training to heterogeneous spatio-temporal data.
\sys{} first embeds each trip’s numerical and linguistic features into a shared latent space, then employs similarity-aware clustering to distribute samples across shards so that future deletions touch only a single constituent model while preserving inter-shard diversity. Each shard is trained incrementally; at inference time, constituent predictions are aggregated to yield the output. Deletion requests trigger retraining solely of the affected shard from its last valid checkpoint, guaranteeing exact unlearning.
Experiments on a ten-month real-world mobility log demonstrate that \sys{} (i) sustains baseline predictive accuracy, and (ii) consistently outperforms random sharding in both error and convergence speed. These results establish MobText-SISA as a practical foundation for privacy-compliant analytics on multimodal mobility data at urban scale.

\color{black}

\end{abstract}


%   <concept>
%       <concept_id>10003033.10003099.10003101</concept_id>
%       <concept_desc>Networks~Location based services</concept_desc>
%       <concept_significance>300</concept_significance> 

%       </concept>
%   <concept>
%       <concept_id>10003120.10003138</concept_id>
%       <concept_desc>Human-centered computing~Ubiquitous and mobile computing</concept_desc>
%       <concept_significance>300</concept_significance>
%       </concept>
%  </ccs2012>

\color{blue}
\color{black}

\maketitle

% \vspace{-0.4cm}
\section{Introduction}
% \color{blue}

The profound demographic shift towards an ageing global population is placing unprecedented strain on health and social-care infrastructures. The World Health Organization (WHO) estimates that 142 million older adults are unable to live independently, with approximately two-thirds expected to require long-term care in their lifetime\cite{WHO:Long-term-care}. This demand is compounded by a projected global shortfall of 11 million health and care workers by 2030, further exacerbating the gap between need and supply\cite{WHO:Health-workforce}.
To help mitigate this imbalance, care-oriented taxi services (Care Taxis) are being deployed worldwide to provide essential mobility assistance. For instance, Neighborly Care Network in Florida\footnote{https://neighborly.org/} facilitates attendance at senior day-care centers, while New Zealand's R\&R Total Mobility\footnote{https://randrmobility.co.nz/} scheme subsidizes travel for older adults and persons with disabilities. Japan has similarly expanded its community-based transport services under its long-term-care insurance system.

The operational workflow of these services generates rich, longitudinal data. Providers dispatch nursing-care taxis, and each vehicle's onboard system logs timestamps for depot departures and arrivals at user residences. During dispatch, operators record the rider’s profile (including age, gender, and mobility attributes like cane use or wheelchair dependence), along with free-text notes detailing special assistance needs. These multimodal logs constitute the dataset analyzed in this study.

Concurrently, the advent of large-scale foundation models is transforming geospatial analytics. Recent advancements like Google's Geospatial Reasoning models and the Trajectory-Powered Foundation Model of Mobility\cite{10.1145/3681766.3699610} promise to unlock insights from location traces for applications in public health and urban planning. However, these mobility traces contain highly sensitive information, such as home addresses and daily routines. This creates significant privacy risks, as machine learning models are vulnerable to attacks like membership inference (MIA)\cite{shokri2017membership} and attribute inference (AIA)\cite{9581166}, which can expose private data from model outputs. The threat is amplified by recent jailbreak techniques that can bypass safety filters in large models, enabling malicious data extraction\cite{deng2024masterkey}.
These concerns have motivated a growing body of research into privacy-preserving learning, including our own prior work. In particular, we explored synthetic data generation with GANs, employing MIA as a privacy evaluation metric\cite{10214915}. We also examined taxi demand prediction in decentralized settings through the framework of decentralized federated learning\cite{ozeki2024privacy}. Furthermore, we also investigated the susceptibility of decision-tree based models that predict pickup times for users of paratransit services to AIA, discussing potential risks of user attribute leakage\cite{Takeuchi2025_UrbComp}.

% Against this backdrop of heightened risk, data protection regulations such as the GDPR have introduced the "Right to be Forgotten." 
These efforts highlight both the promise and the limitations of existing privacy-preserving techniques, motivating stronger guarantees such as the "Right to be Forgotten" enshrined in GDPR.
This principle mandates that individuals can have their personal data erased and, critically, that its influence on trained models be eliminated. Machine unlearning is a computational paradigm designed to address this requirement by efficiently removing a specific user's contribution from a trained model.

% \section{Proposed Method}
\begin{figure*}[t]
    \centering
    \includegraphics[width=0.95\linewidth]{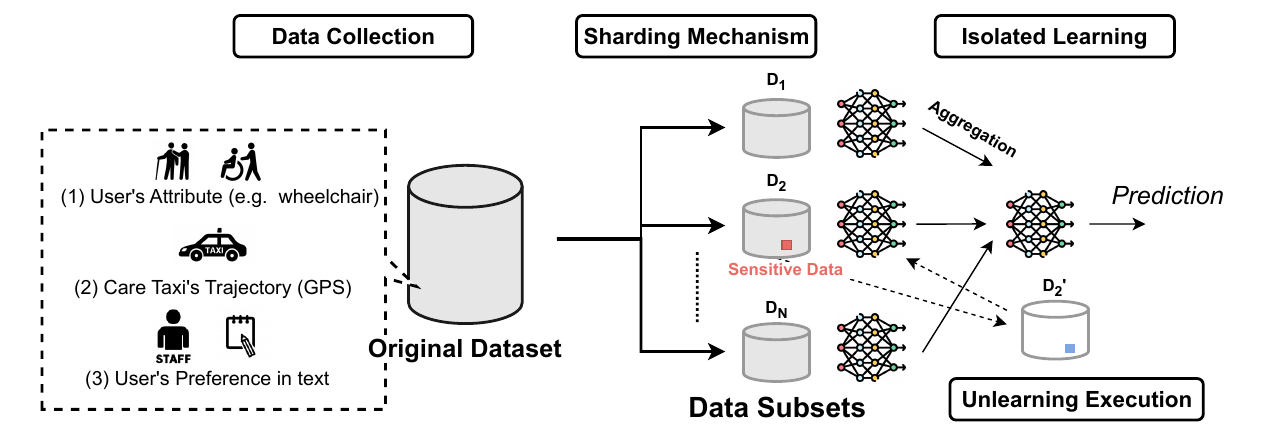}
    % \vspace{-0.4cm}
    \caption{System Overview.}
    \label{fig:sisa_system_overview}
\end{figure*}

% Among existing frameworks, Sharded, Isolated, Sliced, and Aggregated (SISA) training\cite{bourtoule2021machine, 10.1145/3678717.3691269} offers a practical approach. SISA partitions the dataset into disjoint shards, training an independent "constituent" model on each. Each shard is further divided into temporal slices, and the constituent is trained incrementally, with its parameters checkpointed before each new slice is introduced. For inference, the ensemble's predictions are aggregated. When a data point must be unlearned, only the single constituent trained on the slice containing that point is retrained from the last valid checkpoint. This localized process drastically reduces the computational cost compared to full model retraining.
% While SISA offers significant computational advantages for machine unlearning, recent studies have highlighted its adverse effects on model performance, particularly concerning minority classes\cite{koch2023no}. 
% Despite these findings, there remains a lack of research focused on developing shard partitioning methods aimed at preserving overall model performance post-partitioning. Specifically, strategies that proactively address the performance degradation resulting from data partitioning, beyond considerations of class imbalance, are yet to be thoroughly explored.
Among existing machine-unlearning frameworks, Sharded, Isolated, Sliced, and Aggregated (SISA) training \cite{bourtoule2021machine,10.1145/3678717.3691269} offers a pragmatic solution.
SISA first partitions the dataset into mutually exclusive shards and trains an independent constituent model on each shard.
Each shard is then segmented into temporal slices; the constituent model is trained incrementally, with its parameters checkpointed immediately before a new slice is introduced.
During inference, predictions from all constituents are aggregated to yield the ensemble output.
When a data sample must be unlearned, only the constituent whose slice contains that sample is retrained from its last valid checkpoint, drastically reducing computational overhead compared with full-model retraining.
% Despite these efficiency gains, recent studies have shown that SISA can degrade predictive performance, especially for minority classes \cite{koch2023no}. Yet little research has explored shard-partitioning strategies that preserve overall accuracy after partitioning. In particular, proactive methods that mitigate performance degradation arising from data partitioning, beyond merely addressing class imbalance, remain largely unexamined.
Despite these efficiency gains, recent studies show that SISA’s effectiveness hinges on the statistical similarity of the shards: when the data distributions differ markedly, constituent models learn shard-specific boundaries that fail to generalise, so the aggregated ensemble can underperform a monolithic baseline even when class proportions remain balanced\cite{koch2023no}. 
In other words, performance degradation is driven by inter-shard distributional heterogeneity—loss of synergistic information, covariate shift, and feature sparsity—rather than by class imbalance alone.  
Moreover, partitioning inevitably reduces the amount of data available to each learner; small shards encourage underfitting or overfitting to idiosyncratic patterns.
Yet little research has explored shard-partitioning strategies that preserve overall accuracy after partitioning. Proactive methods that explicitly minimise inter-shard divergence while maintaining unlearning locality therefore remain an open—and largely unexplored—research direction.

% Leveraging real-world care-taxi trip logs, comprising location trajectories, mobility attributes, and textual notes, this study empirically investigates the consequences of data deletion on model performance and privacy. We specifically focus on how this inference capability changes after the corresponding user data is unlearned via the SISA framework, and examine how different sharding assignment strategies affect both unlearning efficiency and overall model fidelity.
Leveraging real-world care-taxi trip logs, comprising location trajectories, mobility attributes, and textual notes, this study empirically investigates the consequences of data deletion on model performance and privacy. We specifically focus on how this inference capability changes after the corresponding user data is unlearned via the SISA framework, and examine how different sharding assignment strategies affect both unlearning efficiency and overall model fidelity. To this end, we introduce a similarity-aware shard-assignment strategy: before training, the multimodal records are clustered in a latent space, and each shard is populated with balanced samples drawn from every cluster. By homogenizing the latent data distribution across shards, this clustering-based allocation reduces inter-shard heterogeneity and thereby alleviates the accuracy loss identified in the preceding discussion.

\textit{Our contributions are summarized as follows. First, this is the first study to examine privacy threats and, in particular, deletion effects on machine learning models in the care-taxi domain from the perspective of machine unlearning. We discuss the application of the machine unlearning framework known as SISA and propose methodological extensions to realize machine unlearning efficiently. Second, in the context of machine unlearning, we evaluate practical applicability by handling multi-modal data comprising trip logs and associated textual information. Third, through a ten-month data collection and annotation process in collaboration with care providers, we have obtained a rich care-taxi dataset that can benefit the broader research community. Finally, we cast the empirical study as a supervised task that predicts the end-to-end pickup duration for each trip, leveraging the multimodal attributes collected by care providers.}

\color{black}

\section{Proposed Method}
Figure~\ref{fig:sisa_system_overview} shows the overview of our proposed method. 
This method includes three modules and assumes an unlearning request.%,  \textbf{Data Collection}, \textbf{Sharding Mechanism}, \textbf{Isolated Learning and Aggregation}.
% This data collection leverages a \textbf{Data Recorder}  mobile application, running on each WiFi-enabled device, utilizing the RTT API~\cite{RTT_API} to gather RTT readings. This data is then uploaded to a server for further processing.
\textbf{Data Collection} is the process by which care service providers in Japan record data when picking up users using nursing care taxis.
\textbf{Sharding Mechanism} partitions the collected dataset(Sharded dataset) to enable efficient learning in scenarios where unlearning is required.
\textbf{Isolated Learning and Aggregation} refers to the procedure in which individual machine learning models are trained on each sharded dataset, and their outputs are aggregated to perform prediction with the overall model.

\subsection{Data Collection}
Care providers dispatch nursing-care taxis that follow home–facility and facility–home itineraries.  
For every trip, the onboard system records departure and arrival times as well as the precise geographic coordinates of the journey.
Dispatch operators simultaneously store passenger profiles—age, gender, and mobility indicators such as cane use or wheelchair dependence—and optionally add free-text notes that describe special assistance (for example, “call five minutes before arrival” or “seat in the front passenger seat”).  
\textit{Although such notes are rarely exploited in mobility analytics, we treat them as a first-class input modality and integrate them with the numeric features to enrich the representation of each trip instance.}  
Such information is considered potentially useful for predicting the total duration required for the pickup process.  
We additionally include a naïve travel-time feature, computed by dividing the route length retrieved from OpenStreetMap's routing API\footnote{https://project-osrm.org/} by the applicable speed limit.

% \vspace{-1.3cm}
\subsection{Sharding Mechanism}
% \color{blue}
% In the Sharding Mechanism, the dataset is partitioned into multiple subsets (shards).  
% The core strategy here is to predefine the partitioning in such a way that, when a data unlearning request is issued, the associated computational overhead for re-training can be minimized.  
% We assume that the number of shards, denoted by $k$, is fixed in advance.  
% Our method aims to distribute data samples across these $k$ shards in a way that allocates similar samples to different subsets as evenly as possible.  
% By doing so, we expect to accelerate the convergence of each isolated model during re-training, since training models on internally heterogeneous but label-consistent data can be more efficient than resolving conflicts among similar inputs with differing labels.  
% This is achieved by constructing subsets where the dissimilarity between input samples is maximized, thereby making the classification task within each shard more tractable.  
% Such an approach is also applicable in the context of resource-constrained machine learning~\cite{CACTUS}.
% To determine the similarity between input samples, we apply t-SNE to obtain a low-dimensional representation of the input data.  
% Based on this representation, we construct clusters using the $k$-nearest neighbors (kNN) algorithm.  
% Subsequently, we generate each sub-dataset by iteratively selecting one sample from each cluster, thereby ensuring that similar data are evenly distributed across the subsets.
\begin{figure}
    \centering
    \includegraphics[width=0.95\linewidth]{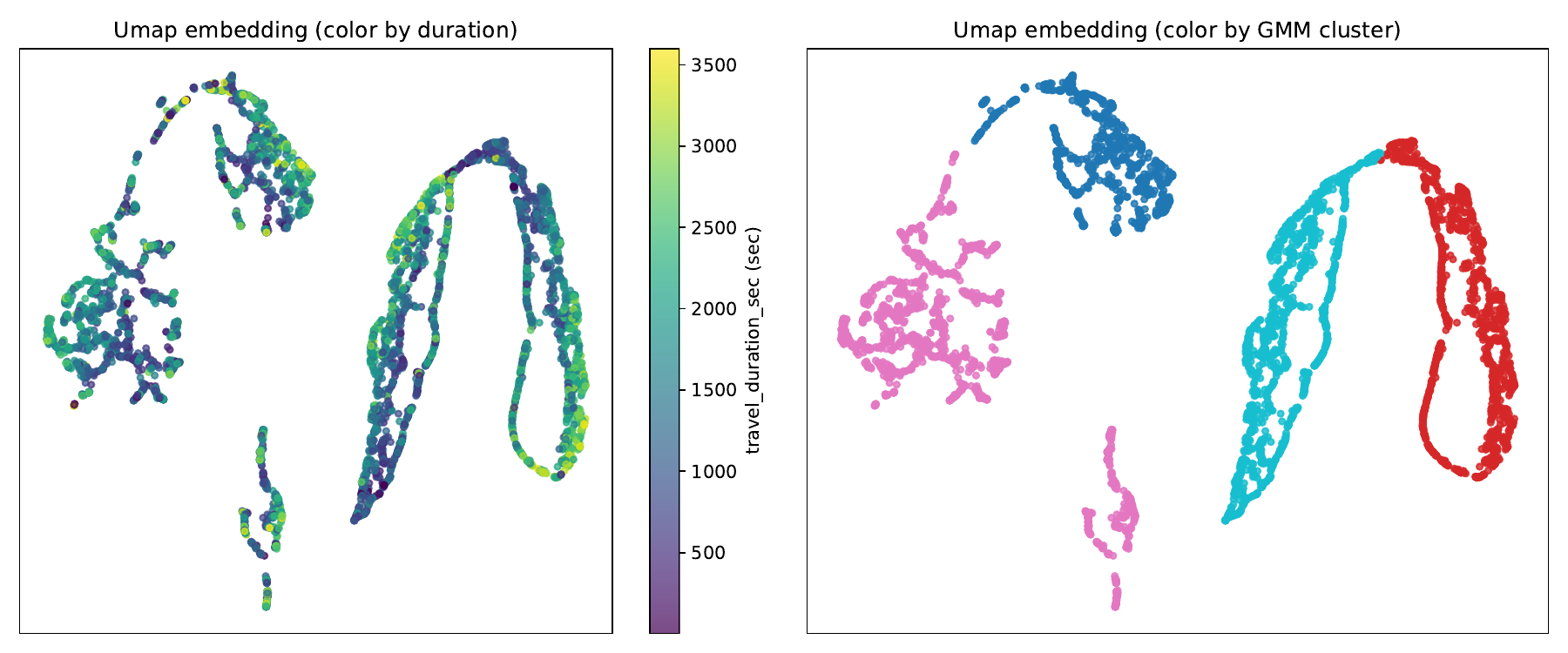}
    % \vspace{-0.4cm}
    \caption{Umap Visualization\cite{becht2019dimensionality} and Gaussian mixture model-based clustering.}
    \label{fig:umap_gmm}
\end{figure}

The dataset is divided into \(k\) disjoint shards that can later be retrained independently when a right-to-be-forgotten request arrives.  
\textit{The partitioning seeks two outcomes: first, any single deletion request should affect as few shards as possible; second, the aggregated predictor should retain high accuracy.}  
Every trip—numeric attributes plus free-text note—is mapped to a unified feature vector.  
The note is embedded with a frozen BERT encoder\cite{devlin2019bert}, then compressed to 32 dimensions via PCA; the result is concatenated with the scaled numeric fields.  
We project these vectors into a two-dimensional manifold with UMAP for visual inspection, as shown in Figure~\ref{fig:umap_gmm}.  
A Gaussian mixture model\cite{reynolds2009gaussian} defined on this space provides a soft similarity metric among trips.  
\textit{Clusters are traversed in a round-robin manner, one sample per shard, so highly similar trips land in distinct shards.}  
This spreads deletion impact while preserving representativeness, echoing the diversity-injection principle of CACTUS\cite{CACTUS}.  
\textit{Because each shard now receives only a fraction of the full dataset, sample scarcity can itself degrade model performance.  Our balanced cluster-aware assignment mitigates this risk by ensuring that no shard is deprived of rare but informative patterns.}

% \subsection{Isolated Learning and Aggregation}
% For each individually constructed sub-dataset, a machine learning model with an identical architecture is trained. 
% To handle text data in a sharded dataset, we employ BERT to get the embeddings of each text. While training, the BERT is frozen, and the MLP is trained to get better vectors.
% Within each model, early stopping is applied to prevent overfitting and to ensure that the training process terminates at an appropriate point while preserving generalization performance. 
% \textcolor{red}{The final output is determined by performing a confidence-weighted majority voting across the predictions of all trained models.}
\subsection{Isolated Learning with Aggregation}
% Each shard feeds an identical model architecture that combines structured and textual inputs.  
% The textual note is embedded with a frozen BERT encoder\cite{devlin2019bert}, and the resulting vector is concatenated with the numeric attributes before being passed to a multilayer perceptron. 
% Before this concatenation, the BERT-derived embedding is further reduced to 32 dimensions via Principal Component Analysis (PCA), allowing for dimensionality reduction while preserving key semantic information.
Each shard feeds an identical model architecture that combines the structured fields with the cluster-aware text embedding described above.  
At inference time, the ensemble prediction is obtained through aggregation among constituent models.
For all shard-specific models, we use a consistent baseline architecture consisting of a three-layer multilayer perceptron (MLP), and early stopping is applied to prevent overfitting.  
% with hidden dimensions [128, 64, 64]. Each hidden layer is followed by Layer Normalization, ReLU activation, and a dropout layer with a rate of 0.2. This design ensures stable training dynamics across shards without relying on batch-level statistics.
The classification task is defined at one-minute intervals, resulting in a large number of possible classes. To handle this fine-grained temporal granularity, we adopt an exponential label smoothing strategy. This approach assigns soft targets based on the exponential decay of distance from the true label in the time axis, enabling the model to learn smoother decision boundaries and tolerate minor deviations in prediction.
\textit{We choose to frame this task as classification rather than regression to improve learning stability. Regression is particularly sensitive to distribution shifts and heteroscedastic noise across clients, which can result in outlier-dominated updates and unstable optimization. In contrast, classification with normalized cross-entropy loss ensures bounded gradients and consistent aggregation, making it more robust under the data.}

% \subsection{Unlearning Execution}
% When performing unlearning, the machine learning model corresponding to the sub-dataset that contains the data subject to deletion is retrained using only the remaining data, excluding the requested samples.  
% The SISA mechanism is categorized as an \textit{exact unlearning} approach, where it is essential that the retraining process is carried out without any use of the data that have been requested for deletion.
\subsection{Unlearning Execution}
When the system receives a right-to-be-forgotten request, it identifies the shard or shards that contain the target records and discards any checkpoints created after those records were incorporated.  
Only the affected constituent models are retrained, using exactly the retained data; all other shards and their models remain unchanged.  
\textit{This procedure achieves exact unlearning while reducing computational cost compared to full retraining.}
In the original SISA framework, this retraining process is further optimized through a technique called Slice, in which each shard is subdivided into multiple sub-datasets. Checkpoints are cached during training so that, upon a deletion request, retraining can resume from an intermediate state rather than starting from scratch.
However, in this study, we do not employ the Slice mechanism or any caching strategy. Instead, we perform full retraining of the affected shard models from the initial state, without relying on any previously saved checkpoints. This allows us to isolate and evaluate the effects of our proposed shard partitioning strategy in a controlled and interpretable setting.

\color{black}

\section{Evaluation}
\label{sec:evaluation}

\color{black}

\subsection{Dataset}
To empirically validate our method, we utilize real-world pick-up and drop-off data provided by a specific care facility, spanning the period from October 1 to October 31, 2023, and consisting of 5193 unique records. 
% The geographical relationship between the facility and its users is visualized in Figure~\ref{fig:user_distributions}.

% \begin{figure}[t]
%     \centering
%     % \includegraphics[width=\linewidth]{imgs/user_distributions.pdf}
%     \includegraphics[width=0.85\linewidth, height=5.5cm]{imgs/user_distribution.png}
%     \caption{Addresses of the facility and users (the background map has been altered to an unrelated location to protect personal information).}
%     \label{fig:user_distributions}
% \end{figure}

\subsection{Evaluation of Prediction Performance}
\begin{figure}
    \centering
    \includegraphics[width=0.95\linewidth]{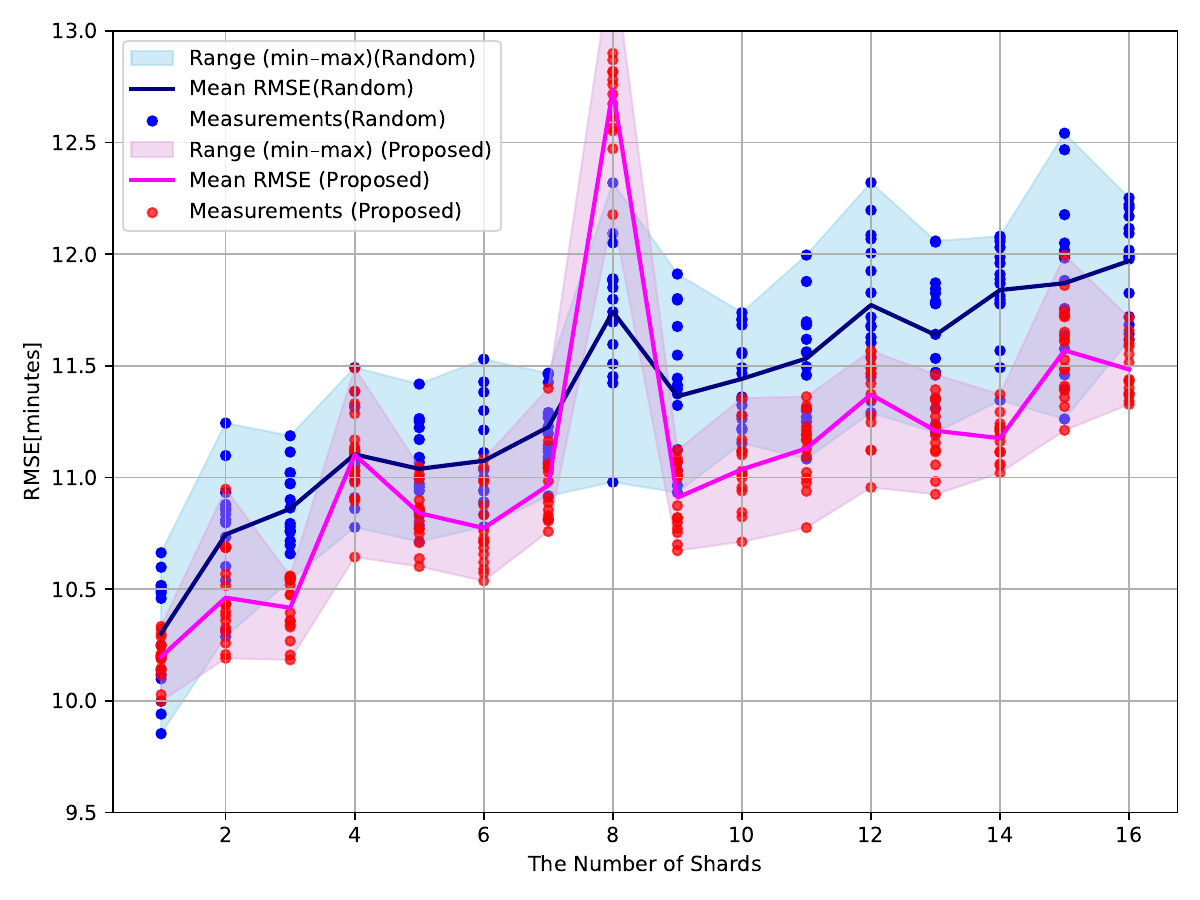}
    % \vspace{-0.4cm}
    \caption{RMSE vs. The Number of Shards.}
    \label{fig:RMSE_vs_The_Number_of_Shards}
\end{figure}

\begin{figure}
    \centering
    \includegraphics[width=0.95\linewidth]{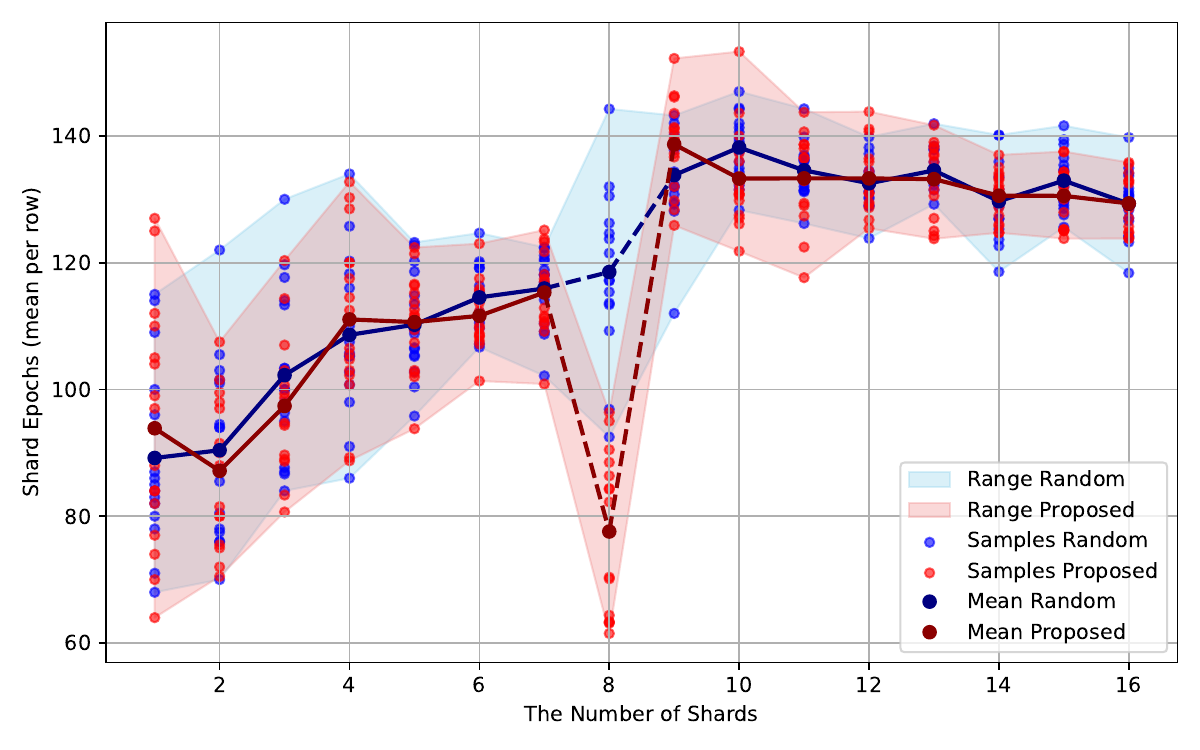}
    % \vspace{-0.4cm}
    \caption{Training Epochs vs. The Number of Shards.}
    \label{fig:epochs_vs_The_Number_of_Shards}
\end{figure}

Figure~\ref{fig:RMSE_vs_The_Number_of_Shards} shows an evaluation of the relationship between the number of shards and the prediction error of the model. The evaluation metric used in this study is the Root Mean Square Error (RMSE), and thus, the unit of error is expressed in minutes.
It can be observed that the proposed method consistently outperforms the model trained on randomly partitioned datasets, regardless of the number of shards. 
% The aggregated model obtained through the proposed partitioning strategy demonstrates superior performance.
A general trend is observed in which the overall prediction performance deteriorates as the number of shards increases.
% This phenomenon can be attributed to the proposed data partitioning method, which intentionally forms shards composed of dissimilar data samples. 
The proposed partitioning strategy mitigates the accuracy degradation that ordinarily accompanies an increasing number of shards by deliberately distributing dissimilar data samples across shards, thereby enabling each constituent model to learn more discriminative and complementary patterns.
% This facilitates the learning of more discriminative patterns by the prediction model, thereby mitigating the negative effects typically caused by data scarcity in individual shards.
% By ensuring that each shard consists of heterogeneous data, the model is encouraged to learn more generalizable representations. 
This facilitates robust classification even under limited data conditions, thereby alleviating the typical performance degradation associated with excessive data fragmentation.
% Figure \ref{fig:RMSE_vs_The_Number_of_Shards} illustrates how prediction error varies with the number of shards. The metric reported is the Root Mean Square Error (RMSE), expressed in minutes. Across every shard count, the ensemble produced by our similarity-aware partitioning yields lower RMSE than the model trained on randomly partitioned data. Although a gradual increase in error is inevitable as the dataset is divided into ever-smaller shards, our strategy markedly restrains this deterioration. By distributing dissimilar trips evenly among shards, the method supplies each learner with a heterogeneous yet representative sample mix, fostering complementary decision boundaries and safeguarding aggregate accuracy even under pronounced data fragmentation.

% \vspace{-0.3cm}
\subsection{Evaluation of Unlearning Scenario}
Figure~\ref{fig:epochs_vs_The_Number_of_Shards} shows the relationship between the number of shards and the average number of training epochs required for each partitioned model. In this evaluation, early stopping was applied based on the training data, enabling us to measure the convergence behavior of each model.
In the context of machine unlearning, the number of training epochs can be interpreted as an indicator of how quickly a model can be retrained following a user's request for data deletion. While the overall difference between the proposed method and random partitioning is not substantial, the average and minimum number of epochs per shard are lower for the proposed method in a majority of cases, specifically, for shard counts ranging from 2 to 16, excluding the exceptional case when the number of shards is 8. This suggests that the proposed partitioning strategy is marginally more efficient than random partitioning in terms of retraining speed under deletion constraints.

% Furthermore, across both Figures~\ref{fig:RMSE_vs_The_Number_of_Shards} and ~\ref{fig:epochs_vs_The_Number_of_Shards}, the case with 8 shards presents as an outlier in the results of the proposed method. This anomaly is likely due to the model prematurely converging to a local optimum, resulting in early termination of training.
Furthermore, across both Figures \ref{fig:RMSE_vs_The_Number_of_Shards} and \ref{fig:epochs_vs_The_Number_of_Shards}, the case with the 8-shard setting emerges as an outlier for the proposed method. This anomaly is likely attributable to two intertwined factors: (i) premature convergence of the constituent models to local optima, which triggers early termination of training, and (ii) the possibility that the Gaussian-mixture-model clustering failed to yield coherent clusters at this particular shard count, thereby degrading the subsequent partitioning quality.

% \vspace{-0.3cm}
\section{Conclusion}
In this study, we investigated a data shard partitioning strategy to enable efficient machine unlearning on mobility data containing linguistic information, using the SISA framework. Our method performs clustering on the two-dimensional coordinates obtained by projecting compressed input representations, via Uniform Manifold Approximation and Projection (UMAP), followed by Gaussian Mixture Modeling (GMM). The resulting clusters are then distributed across shards in a round-robin manner. This approach preserves the underlying data distribution within each shard, allowing models trained on reduced datasets to maintain predictive accuracy without significant degradation, even after partitioning.

While this work focuses on the data partitioning methodology, further research is warranted regarding the aggregation of machine learning models trained independently on each shard. Effective aggregation techniques may enhance both the performance and robustness of the overall system under deletion and retraining constraints.

% \vspace{-0.15cm}
% \begin{acks}
% \vspace{-0.15cm}
% This work was supported by JST, CREST Grant JPMJCR21M5, JST
% PRESTO Grant JPMJPR2361, and JST BOOST, Japan Grant Number JPMJBS2402.
% \end{acks}

% \vspace{-0.3cm}
\bibliographystyle{ACM-Reference-Format}
\bibliography{sample}


\begin{thebibliography}{16}

%%% the Web address.

\ifx \showCODEN    \undefined \def \showCODEN     #1{\unskip}     \fi
\ifx \showDOI      \undefined \def \showDOI       #1{#1}\fi
\ifx \showISBNx    \undefined \def \showISBNx     #1{\unskip}     \fi
\ifx \showISBNxiii \undefined \def \showISBNxiii  #1{\unskip}     \fi
\ifx \showISSN     \undefined \def \showISSN      #1{\unskip}     \fi
\ifx \showLCCN     \undefined \def \showLCCN      #1{\unskip}     \fi
\ifx \shownote     \undefined \def \shownote      #1{#1}          \fi
\ifx \showarticletitle \undefined \def \showarticletitle #1{#1}   \fi
\ifx \showURL      \undefined \def \showURL       {\relax}        \fi
% invisible to TeX
\providecommand\bibfield[2]{#2}
\providecommand\bibinfo[2]{#2}
\providecommand\natexlab[1]{#1}
\providecommand\showeprint[2][]{arXiv:#2}

\bibitem[\protect\citeauthoryear{Becht et~al\mbox{.}}{Becht et~al\mbox{.}}{2019}]%
        {becht2019dimensionality}
\bibfield{author}{\bibinfo{person}{Etienne Becht} {et~al\mbox{.}}} \bibinfo{year}{2019}\natexlab{}.
\newblock \showarticletitle{Dimensionality reduction for visualizing single-cell data using UMAP}.
\newblock \bibinfo{journal}{\emph{Nature biotechnology}} \bibinfo{volume}{37}, \bibinfo{number}{1} (\bibinfo{year}{2019}), \bibinfo{pages}{38--44}.
\newblock


\bibitem[\protect\citeauthoryear{Bourtoule et~al\mbox{.}}{Bourtoule et~al\mbox{.}}{2021}]%
        {bourtoule2021machine}
\bibfield{author}{\bibinfo{person}{Lucas Bourtoule} {et~al\mbox{.}}} \bibinfo{year}{2021}\natexlab{}.
\newblock \showarticletitle{Machine unlearning}. In \bibinfo{booktitle}{\emph{2021 IEEE symposium on security and privacy (SP)}}. IEEE, \bibinfo{pages}{141--159}.
\newblock


\bibitem[\protect\citeauthoryear{Choudhury et~al\mbox{.}}{Choudhury et~al\mbox{.}}{2024}]%
        {10.1145/3681766.3699610}
\bibfield{author}{\bibinfo{person}{Shushman Choudhury} {et~al\mbox{.}}} \bibinfo{year}{2024}\natexlab{}.
\newblock \showarticletitle{Towards a Trajectory-powered Foundation Model of Mobility}. In \bibinfo{booktitle}{\emph{Proceedings of the 3rd ACM SIGSPATIAL International Workshop on Spatial Big Data and AI for Industrial Applications}} \emph{(\bibinfo{series}{GeoIndustry '24})}. \bibinfo{publisher}{Association for Computing Machinery}, \bibinfo{address}{New York, NY, USA}, \bibinfo{pages}{1^^e2^^80^^934}.
\newblock
\showISBNx{9798400711459}


\bibitem[\protect\citeauthoryear{Deng et~al\mbox{.}}{Deng et~al\mbox{.}}{2024}]%
        {deng2024masterkey}
\bibfield{author}{\bibinfo{person}{Gelei Deng} {et~al\mbox{.}}} \bibinfo{year}{2024}\natexlab{}.
\newblock \showarticletitle{MASTERKEY: Automated Jailbreaking of Large Language Model Chatbots}. In \bibinfo{booktitle}{\emph{NDSS}}.
\newblock


\bibitem[\protect\citeauthoryear{Devlin et~al\mbox{.}}{Devlin et~al\mbox{.}}{2019}]%
        {devlin2019bert}
\bibfield{author}{\bibinfo{person}{Jacob Devlin} {et~al\mbox{.}}} \bibinfo{year}{2019}\natexlab{}.
\newblock \showarticletitle{Bert: Pre-training of deep bidirectional transformers for language understanding}. In \bibinfo{booktitle}{\emph{Proceedings of the 2019 conference of the North American chapter of the association for computational linguistics: human language technologies, volume 1 (long and short papers)}}. \bibinfo{pages}{4171--4186}.
\newblock


\bibitem[\protect\citeauthoryear{Kanza, Krishnamurthy, and Srivastava}{Kanza et~al\mbox{.}}{2024}]%
        {10.1145/3678717.3691269}
\bibfield{author}{\bibinfo{person}{Yaron Kanza}, \bibinfo{person}{Balachander Krishnamurthy}, {and} \bibinfo{person}{Divesh Srivastava}.} \bibinfo{year}{2024}\natexlab{}.
\newblock \showarticletitle{A Geospatial Perspective on Data Ownership, the Right to be Forgotten, Copyrights, and Plagiarism in Generative AI}. In \bibinfo{booktitle}{\emph{Proceedings of the 32nd ACM International Conference on Advances in Geographic Information Systems}} \emph{(\bibinfo{series}{SIGSPATIAL '24})}. \bibinfo{publisher}{Association for Computing Machinery}, \bibinfo{address}{New York, NY, USA}, \bibinfo{pages}{477^^e2^^80^^93480}.
\newblock
\showISBNx{9798400711077}


\bibitem[\protect\citeauthoryear{Koch and Soll}{Koch and Soll}{2023}]%
        {koch2023no}
\bibfield{author}{\bibinfo{person}{Korbinian Koch} {and} \bibinfo{person}{Marcus Soll}.} \bibinfo{year}{2023}\natexlab{}.
\newblock \showarticletitle{No matter how you slice it: Machine unlearning with sisa comes at the expense of minority classes}. In \bibinfo{booktitle}{\emph{2023 IEEE Conference on Secure and Trustworthy Machine Learning (SaTML)}}. IEEE, \bibinfo{pages}{622--637}.
\newblock


\bibitem[\protect\citeauthoryear{Ozeki, Yonekura, Rizk, and Yamaguchi}{Ozeki et~al\mbox{.}}{2023}]%
        {10214915}
\bibfield{author}{\bibinfo{person}{Ren Ozeki}, \bibinfo{person}{Haruki Yonekura}, \bibinfo{person}{Hamada Rizk}, {and} \bibinfo{person}{Hirozumi Yamaguchi}.} \bibinfo{year}{2023}\natexlab{}.
\newblock \showarticletitle{Balancing Privacy and Utility of Spatio-Temporal Data for Taxi-Demand Prediction}. In \bibinfo{booktitle}{\emph{2023 24th IEEE International Conference on Mobile Data Management (MDM)}}. \bibinfo{pages}{215--220}.
\newblock
\urldef\tempurl%
\url{https://doi.org/10.1109/MDM58254.2023.00044}
\showDOI{\tempurl}


\bibitem[\protect\citeauthoryear{Ozeki, Yonekura, Rizk, and Yamaguchi}{Ozeki et~al\mbox{.}}{2024}]%
        {ozeki2024privacy}
\bibfield{author}{\bibinfo{person}{Ren Ozeki}, \bibinfo{person}{Haruki Yonekura}, \bibinfo{person}{Hamada Rizk}, {and} \bibinfo{person}{Hirozumi Yamaguchi}.} \bibinfo{year}{2024}\natexlab{}.
\newblock \showarticletitle{Privacy Preserved Taxi Demand Prediction System for Distributed Data}. In \bibinfo{booktitle}{\emph{Proceedings of the 32nd ACM International Conference on Advances in Geographic Information Systems}}. \bibinfo{pages}{123--134}.
\newblock


\bibitem[\protect\citeauthoryear{Rastikerdar et~al\mbox{.}}{Rastikerdar et~al\mbox{.}}{2024}]%
        {CACTUS}
\bibfield{author}{\bibinfo{person}{Mohammad~Mehdi Rastikerdar} {et~al\mbox{.}}} \bibinfo{year}{2024}\natexlab{}.
\newblock \showarticletitle{CACTUS: Dynamically Switchable Context-aware micro-Classifiers for Efficient IoT Inference}. In \bibinfo{booktitle}{\emph{Proceedings of the 22nd Annual International Conference on Mobile Systems, Applications and Services}} \emph{(\bibinfo{series}{MOBISYS '24})}. \bibinfo{publisher}{ACM}, \bibinfo{address}{New York, NY, USA}, \bibinfo{pages}{505^^e2^^80^^93518}.
\newblock
\showISBNx{9798400705816}


\bibitem[\protect\citeauthoryear{Reynolds}{Reynolds}{2009}]%
        {reynolds2009gaussian}
\bibfield{author}{\bibinfo{person}{Douglas Reynolds}.} \bibinfo{year}{2009}\natexlab{}.
\newblock \showarticletitle{Gaussian mixture models}.
\newblock In \bibinfo{booktitle}{\emph{Encyclopedia of biometrics}}. \bibinfo{publisher}{Springer}, \bibinfo{pages}{659--663}.
\newblock


\bibitem[\protect\citeauthoryear{Shokri, Stronati, Song, and Shmatikov}{Shokri et~al\mbox{.}}{2017}]%
        {shokri2017membership}
\bibfield{author}{\bibinfo{person}{Reza Shokri}, \bibinfo{person}{Marco Stronati}, \bibinfo{person}{Congzheng Song}, {and} \bibinfo{person}{Vitaly Shmatikov}.} \bibinfo{year}{2017}\natexlab{}.
\newblock \showarticletitle{Membership inference attacks against machine learning models}. In \bibinfo{booktitle}{\emph{2017 IEEE Symposium on Security and Privacy (SP)}}. IEEE, \bibinfo{pages}{3--18}.
\newblock


\bibitem[\protect\citeauthoryear{Takeuchi, Yonekura, Yamashita, and Yamaguchi}{Takeuchi et~al\mbox{.}}{2025}]%
        {Takeuchi2025_UrbComp}
\bibfield{author}{\bibinfo{person}{Yuya Takeuchi}, \bibinfo{person}{Haruki Yonekura}, \bibinfo{person}{Kyosuke Yamashita}, {and} \bibinfo{person}{Hirozumi Yamaguchi}.} \bibinfo{year}{2025}\natexlab{}.
\newblock \showarticletitle{Evaluating Attribute Inference Risks in Urban Care Taxi Arrival Time Prediction Models Using Geospatial Data}. In \bibinfo{booktitle}{\emph{Proceedings of the 14th International Workshop on Urban Computing (UrbComp)}} \emph{(\bibinfo{series}{in conjunction with the 31st ACM International Conference on Knowledge Discovery and Data Mining(SIGKDD)})}.
\newblock


\bibitem[\protect\citeauthoryear{{World Health Organization}}{{World Health Organization}}{2022}]%
        {WHO:Long-term-care}
\bibfield{author}{\bibinfo{person}{{World Health Organization}}.} \bibinfo{year}{2022}\natexlab{}.
\newblock \bibinfo{title}{{Long-term care}}.
\newblock \bibinfo{howpublished}{\url{https://www.who.int/europe/news-room/questions-and-answers/item/long-term-care}}.
\newblock
\newblock
\shownote{Accessed: May 15, 2025.}


\bibitem[\protect\citeauthoryear{{World Health Organization}}{{World Health Organization}}{2025}]%
        {WHO:Health-workforce}
\bibfield{author}{\bibinfo{person}{{World Health Organization}}.} \bibinfo{year}{2025}\natexlab{}.
\newblock \bibinfo{title}{{Health workforce}}.
\newblock \bibinfo{howpublished}{\url{https://www.who.int/health-topics/health-workforce}}.
\newblock
\newblock
\shownote{Accessed: May 15, 2025.}


\bibitem[\protect\citeauthoryear{Zhao et~al\mbox{.}}{Zhao et~al\mbox{.}}{2021}]%
        {9581166}
\bibfield{author}{\bibinfo{person}{Benjamin Zi~Hao Zhao} {et~al\mbox{.}}} \bibinfo{year}{2021}\natexlab{}.
\newblock \showarticletitle{{ On the (In)Feasibility of Attribute Inference Attacks on Machine Learning Models }}. In \bibinfo{booktitle}{\emph{2021 IEEE European Symposium on Security and Privacy (EuroS{\&}P)}}. \bibinfo{publisher}{IEEE Computer Society}, \bibinfo{address}{Los Alamitos, CA, USA}, \bibinfo{pages}{232--251}.
\newblock


\end{thebibliography}

\end{document}